\documentclass{article} % For LaTeX2e
\usepackage[final]{colm2026_conference}
\usepackage[export]{adjustbox}
\usepackage{microtype}
\usepackage{hyperref}
\usepackage{url}
\usepackage{booktabs}

\usepackage{lineno}

\definecolor{darkblue}{rgb}{0, 0, 0.5}
\hypersetup{colorlinks=true, citecolor=darkblue, linkcolor=darkblue, urlcolor=darkblue}

\usepackage{multirow}
\usepackage{pdfpages}
\usepackage{afterpage}
\usepackage{stfloats}
\usepackage{xspace}
\usepackage[normalem]{ulem}
\useunder{\uline}{\ul}{}
\usepackage{enumitem}
\usepackage[table]{xcolor}
\usepackage{colortbl}
\usepackage{makecell}
\usepackage{xfp} 

\usepackage{algorithmic}
\usepackage[T1]{fontenc}
\usepackage{tcolorbox}
\usepackage{listings}
\tcbuselibrary{breakable}

\newcommand{\ModelName}{\textbf{\textcolor{darkgray}{\textsc{SEA}}}}

\usepackage{todonotes}

\definecolor{groupgray}{gray}{0.94}

\usepackage[table]{xcolor} % in preamble

\definecolor{lightred}{RGB}{255,230,230}
\definecolor{midred}{RGB}{255,180,180}
\definecolor{darkred}{RGB}{255,120,120}

\usepackage{amsmath}

\title{Joint Optimization of Reasoning and Dual-Memory \\ for Self-Learning Diagnostic Agent}

\author{Bingxuan Li\textsuperscript{$1$}, Simo Du\textsuperscript{$2$}, Yue Guo\textsuperscript{$1$} \\
University of Illinois Urbana-Champaign\textsuperscript{$1$}, Albert Einstein College of Medicine\textsuperscript{$2$} \\
\texttt{bl61@illinois.edu, yueg@illinois.edu}
}

\usepackage{booktabs}
\usepackage{tabularx}
\usepackage{array}

\begin{document}

\ifcolmsubmission
\linenumbers
\fi

\maketitle

\begin{abstract}
Clinical expertise improves not only by acquiring medical knowledge, but by accumulating experience that yields reusable diagnostic patterns. Recent LLMs-based diagnostic agents have shown promising progress in clinical reasoning for decision support. However, most approaches treat cases independently, limiting experience reuse and continual adaptation.
We propose \ModelName{}, a self-learning diagnostic agent with cognitively inspired dual-memory module. We design a reinforcement training framework tailored to our designed agent for joint optimization of reasoning and memory management. We evaluate \ModelName{} in two complementary settings. On standard evaluation with MedCaseReasoning dataset, \ModelName{} achieves 92.46\% accuracy, outperforming the strongest baseline by +19.6\%, demonstrating the benefit of jointly optimizing reasoning and memory. On the long-horizon with ER-Reason dataset, \ModelName{} attains the best final accuracy (0.7214) and the largest improvement (+0.35 $\Delta$Acc@100), while baseline methods show limited or unstable gains. Expert evaluation further indicates that the consolidated rules from \ModelName{} shows strong clinical correctness, usefulness and trust, suggesting that the induced rules in dual-memory module are reliable and practically meaningful. Overall, \ModelName{} improves both diagnostic reasoning ability and continual learning by effectively transforming experience into reusable knowledge.
\end{abstract}

\section{Introduction}
\label{sec:introduction}

In real-world clinical practice, expert diagnostic decision-making does not arise from the accumulation of declarative medical knowledge alone. Instead, it develops through experience-shaped cognitive structures: Compact mental representations that organize knowledge into usable patterns \citep{schmidt1990cognitive,dornan2019experience}. During clinical encounters, clinicians rely on two complementary resources: (i) formal professional knowledge (e.g., disease definitions, guidelines, and pathophysiology) and (ii) experiential reasoning grounded in prior cases, where remembered contexts and outcomes guide the interpretation of new evidence. The latter form of learning is particularly critical for rare diseases and atypical presentations, where textbook descriptions are often insufficient and diagnostically relevant signals are subtle.

Large Language Models (LLMs) have recently shown strong potential for automating clinical reasoning and diagnosis. One line of work frames diagnosis as an \emph{agentic} workflow, emphasizing explicit intermediate steps to enable grounded and auditable reasoning \citep{zhao2025agentic,wang2025medagent,zang2025medical,peng2025tree}. Another line investigates \emph{multi-agent} coordination, such as role specialization, debate, or consensus mechanisms, to improve robustness and efficiency \citep{chen2024rareagents,zuo2025kg4diagnosis,zhao2025confagents,li2025macd,han2025multi,zheng2025end}. Separately, several studies explore long-term learning for agents \citep{qiu2025evolving,wang2021lifelong,10.1145/3442381.3449795,liu2026evomdt}. However, most prior work focuses on teaching agents \emph{what} to diagnose (i.e., acquiring medical knowledge) rather than \emph{how} to improve through accumulated experience, as clinicians do. This gap motivates our central question: \textit{Can agents benefit from experience-driven long-term learning, and how should such experience be stored and reused?}

A key bottleneck is long-term memory management. While general-purpose memory mechanisms have shown promise for agentic AI in web and sandbox environments \citep{peng2025tree,zang2025medical,chen2024rareagents,zuo2025kg4diagnosis}, the medical domain poses distinct challenges: clinical cases are information-dense, evidence is noisy or incomplete, and useful learning often requires abstraction (e.g., reusable heuristics) rather than verbatim storage of trajectories. Na\"{\i}vely retaining every past case quickly becomes infeasible, whereas overly aggressive compression risks discarding rare but diagnostically decisive signals.

To address this issue, we present \ModelName{}, a self-learning diagnostic agent for long-horizon improvement from experience, built around two coupled designs:
First, we introduce a \textbf{dual-memory architecture} inspired by human cognition: a \emph{short-term memory} stores a bounded set of recent, concrete cases, while a \emph{long-term memory} stores abstract diagnostic rules distilled from experience. The agent maintains up to \(K\) case memories in an append-only manner; once this capacity exceeded, selected cases are evicted and summarized into reusable reasoning rules, which are consolidated into long-term memory. Second, we propose a \textbf{reinforcement learning framework that jointly optimize diagnostic reasoning and memory management.} We design structured rollouts that produce traceable reasoning trajectories leading to final diagnosis while explicitly deciding how memory is updated. We then introduce \emph{adaptive rewards} that jointly optimize (i) diagnostic accuracy and (ii) effective memory-management decisions, encouraging continual improvement from experience rather than passive accumulating knowledge.

We evaluate \ModelName{} in two complementary settings: (1) In the standard evaluation on MedCaseReasoning dataset~\citep{wu2025medcasereasoning}, which measures static diagnostic competence without cross-case adaptation, \ModelName{} achieves substantial gains over all baselines, reaching up to 92.46\% accuracy and surpassing the strongest baseline by +19.6\%, demonstrating that jointly optimizing reasoning and memory yields significantly stronger clinical decision-making than outcome-only supervision or naive memory integration. (2) In the long-horizon setting on ER-Reason dataset~\citep{mehandru2025er}, which simulates streaming deployment with feedback, \ModelName{} achieves the best final accuracy of 0.7214 and the largest improvement of +0.35 ($\Delta$Acc@100), while standard and RL-trained models show only marginal gains (e.g., +0.03) or even degradation (e.g., --0.05 at early stages). These results highlight that effective adaptation requires explicit mechanisms for experience accumulation and reuse rather than relying on training alone. Additionally, expert evaluation on the consolidated rules from \ModelName{} yields strong scores in overall trust (4.216), rule usefulness (3.946), and clinical correctness (3.865)on a 5-point Likert scale (5 denotes the highest rating, with an overall average of 3.655, indicating that the induced rules are reliable and practically meaningful. Overall, \ModelName{} not only improves diagnostic reasoning capability but also enables sustained learning during deployment by transforming experience into generalizable knowledge.

\section{Related Work}
\label{sec:related_work}
\noindent\textbf{LLM Agents for Disease Diagnosis.}
Medical foundation models achieve strong performance in biomedical QA and clinical reasoning, indicating that generalist LLMs can serve as clinical decision backbones. \citet{med-palm,singhal2025toward} demonstrate near expert-level QA while highlighting safety gaps. Domain-specialized multimodal models extend this capability to heterogeneous medical data \citep{medgemma,healthgpt}. Concurrently, diagnosis is increasingly framed as an \emph{agentic} process with structured workflows and verifiable steps, including traceable rare-disease systems \citep{zhao2025agentic}, multimodal pipelines \citep{wang2025medagent}, and hierarchical reasoning frameworks \citep{zang2025medical,peng2025tree}. Multi-agent approaches further improve robustness and efficiency via role decomposition and collaboration \citep{chen2024rareagents,zuo2025kg4diagnosis,zhao2025confagents,li2025macd,han2025multi}. Beyond diagnosis, agentic systems also support modular biomedical discovery workflows \citep{ock2025large}.

\noindent\textbf{Memory Mechanisms for Agentic AI.}
A central challenge in clinical agents is maintaining evolving context across multi-step reasoning. Existing systems often rely on structured intermediates (e.g., evidence or logic trees) to track hypotheses \citep{peng2025tree,zang2025medical,chen2024rareagents,zuo2025kg4diagnosis}. More recent work treats memory as an explicit, learnable component that accumulates experience, tracks decisions, and improves future behavior. For example, self-learned knowledge in multi-agent diagnosis acts as persistent memory \citep{li2025macd}, while experience-driven agents use memory to store feedback and adapt over time \citep{yang2025learning}. These trends motivate memory designs that record outcomes, retrieve relevant experience, and refine diagnostic strategies beyond static context windows.

\noindent\textbf{Reinforcement Learning for Long Horizon Self-Evolving.}
Reinforcement learning is increasingly used to enhance long-horizon agent reasoning through iterative feedback. In medicine, RL-based frameworks model diagnosis as sequential hypothesis testing and evidence acquisition \citep{bani2025language}, while clinical agents leverage experiential learning to refine decision policies \citep{lai2025doctor}. More broadly, experience-based RL methods improve both reasoning correctness and process quality over extended trajectories \citep{zhan2025exgrpo,shao2025dr}. Together, these works suggest a unified paradigm for self-evolving agents: combine structured reasoning, explicit memory, and RL signals to enable continual adaptation from experience \citep{yang2025learning}.
\section{Method}
\label{sec:method}

We propose a reinforcement learning framework that jointly optimizes (i) diagnostic reasoning and (ii) memory management for continual
on-the-job improvement.

\begin{figure*}[bt]
    \centering
    \includegraphics[width=1\linewidth]{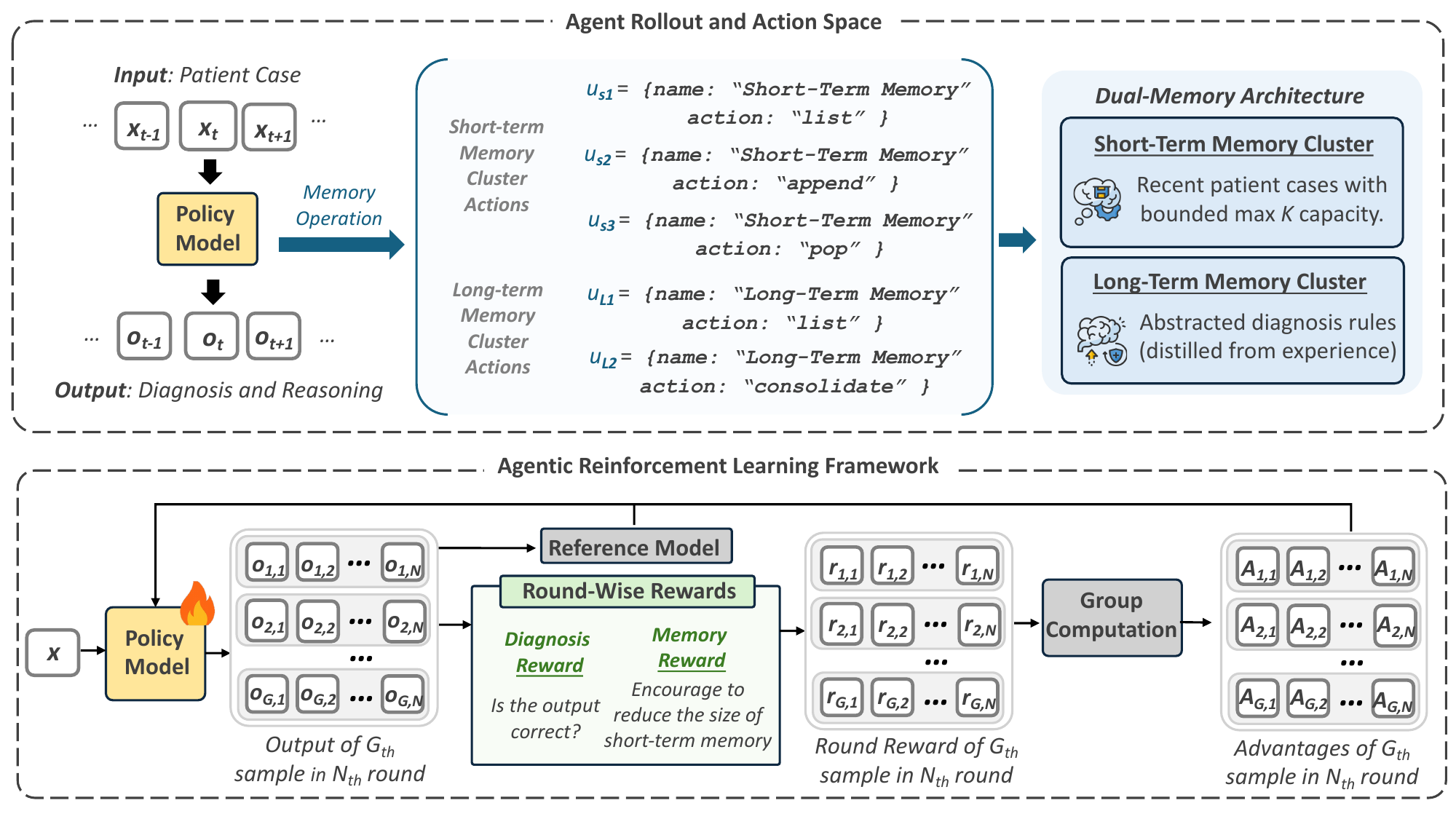}
    \vspace{-0.2in}
    \caption{Overview of \ModelName{}:
At each round $t$, the policy model observes a patient case $x_t$ and may invoke \emph{memory operations} before emitting the final output $o_t$ (diagnosis and reasoning). The agent controls a short-term memory cluster that stores recent patient cases with a bounded capacity $K$ (\texttt{list}/\texttt{append}/\texttt{pop}) and a long-term memory cluster that consolidates experience into abstracted diagnosis rules (\texttt{list}/\texttt{consolidate}). 
For each round, we sample a group of candidate outputs, score them with round-wise rewards, compute group-wise advantages, and update the policy.}

\label{fig:method}
\end{figure*}

\subsection{Agent Design}
\label{sec:method_infra}

\noindent\textbf{Dual-memory architecture.}
A clinician improves by remembering concrete failures/successes and by distilling reusable rules from them.
We want the same behavior from the agent: keep a small set of recent outcome-annotated cases for grounding, while gradually building a compact
rule set that transfers across future patients.
This design also makes the bottleneck explicit: the agent must decide \emph{what to keep} and \emph{what to abstract}, instead of relying on an
external heuristic that may discard the wrong evidence.
Motivated by this, we introduce a dual-memory design inspired by human cognition.
The agent maintains:
(i) a \emph{short-term case memory cluster} $\mathcal{M}^{\text{S}}_t$ that stores a bounded set of recent, concrete cases with outcomes, and
(ii) a \emph{long-term rule memory cluster} $\mathcal{M}^{\text{L}}_t$ that stores abstract diagnostic rules distilled from experience.
The agent state is $s_t = (\mathcal{M}^{\text{S}}_t, \mathcal{M}^{\text{L}}_t)$.

\noindent\textbf{Short-term case memory cluster.}
Recent cases provide high-fidelity evidence (what was observed, what was predicted, and what feedback was returned), which is crucial for
correcting near-miss mistakes and adapting to distribution shifts.
But this evidence is expensive to store in-context, so we enforce a hard capacity $K$ to reflect realistic deployment constraints:
\begin{equation}
\mathcal{M}^{\text{S}}_t = \{c_1, \dots, c_{|\mathcal{M}^{\text{S}}_t|}\}, \qquad |\mathcal{M}^{\text{S}}_t| \le K,
\end{equation}
where each record $c$ contains compact information such as $(x, \mathcal{Y}, \hat{y}, f)$ and optionally a brief rationale snippet.
We use an append-only interface: after finishing round $t$, the new case record is appended, and the agent may trigger a \emph{pop-and-summarize}
operation when the cluster approaches the capacity limit.

\noindent\textbf{Long-term rule memory cluster.}
The long-term memory stores a set of reusable diagnostic rules (e.g., symptom--disease associations, discriminative cues, and failure patterns)
that are distilled from evicted short-term cases:
\begin{equation}
\mathcal{M}^{\text{L}}_t = \{r_1, \dots, r_{|\mathcal{M}^{\text{L}}_t|}\},
\end{equation}
where each rule $r$ is a concise, natural-language (or structured) statement that can be retrieved and applied in later rounds.

\noindent\textbf{Memory operations as agent actions.}
Beyond emitting the diagnostic output, the agent also chooses a memory operation $u_t$ from a small action space:
\begin{equation}
u_t \in \{\texttt{list}, \texttt{append}, \texttt{pop}(p), \texttt{consolidate}\}.
\end{equation}
Intuitively, \texttt{list} retrieves the current contents of the episodic case cluster (and the rules in long-term memory) to support case-based reasoning; \texttt{append} inserts the newly observed case--outcome pair into the case cluster; \texttt{pop}$(p)$ selects one or more cases to evict when the cluster approaches its capacity; and \texttt{consolidate} summarizes the evicted cases into compact diagnostic rules and inserts them into $\mathcal{M}^{\text{L}}$.
Importantly, while the case cluster has a hard capacity $K$, we let the agent \emph{decide when} to pop cases and \emph{how} to distill them into reusable rules.

\noindent\textbf{Structured rollout with Dual-Memory.}
Separating diagnose and memory management makes the agent’s behavior auditable. We can inspect not only what it predicted, but also what it chose
to store as experience and what rules it distilled.
A round-$t$ rollout therefore consists of a diagnostic output and an explicit memory-update decision:
\[
(o_t, u_t) \sim \pi_\theta(\cdot \mid x_t, \mathcal{Y}_t, \mathcal{M}^{\text{S}}_t, \mathcal{M}^{\text{L}}_t),
\]
followed by deterministic state transitions that update $\mathcal{M}^{\text{S}}_{t+1}, \mathcal{M}^{\text{L}}_{t+1}$ according to $u_t$.

\subsection{Reward Modeling}
\label{sec:method_reward}

We design a reward function that encourages the agent to (i) predict correctly and (ii) manage memory efficiently so that it improves over time.
Concretely, we decompose the per-round reward into a \emph{diagnostic reward} and a \emph{memory reward}:
\[
r_t \;=\; \lambda_{\text{diag}} \, r_t^{\text{diag}} \;+\; \lambda_{\text{mem}} \, r_t^{\text{mem}},
\]
where $\lambda_{\text{diag}}, \lambda_{\text{mem}}$ balance diagnostic performance and memory behavior.

\noindent\textbf{Diagnostic reward.}
We use feedback-conditioned rewards that provide strong supervision for correct predictions:
\[
r_t^{\text{diag}} =
\begin{cases}
+5, & \text{if } \hat{y}_t \text{ is correct under } f_t,\\
-5, & \text{otherwise.}
\end{cases}
\]
Here, we set the reward magnitude to $\pm 5$ to balance signal strength and training stability in practice.

\noindent\textbf{Memory-management reward.}
To prevent uncontrolled growth of short-term memory and incentivize timely consolidation, we penalize large cluster occupancy:
\[
r_t^{\text{mem}} \;=\; -\alpha \cdot \frac{|\mathcal{M}^{\text{S}}_t|}{K},
\]
with $\alpha$ controlling the strength ($\alpha=3$ in our implementation). This reward encourages the agent to pop and summarize when appropriate, rather than passively accumulating cases until truncation.

\noindent\textbf{Final per-round reward.}
Putting everything together, the final reward is designed as
\vspace{-0.05in}
\[
r_t \;=\; \lambda_{\text{diag}} \, r_t^{\text{diag}} \;+\; \lambda_{\text{mem}} \, r_t^{\text{mem}}
\]

\subsection{Round-Wise Rewarding}
\label{sec:method_roundwise}
Long-horizon diagnosis exhibits a cold-start challenge: early in the stream, the agent has limited experience for the target population
and its memory is poorly calibrated.
In this phase, directly optimizing only correctness can be high-variance; instead, the most useful behavior is to \emph{acquire and consolidate}
informative case evidence into persistent memory.
As interaction progresses, the agent's memory becomes richer and more stable, and the learning signal should shift toward diagnostic accuracy.

\noindent\textbf{Round-dependent reward shaping.}
To encourage this staged learning, we schedule the trade-off between diagnostic reward and memory reward across rounds.
Let the normalized round index be $i_t = \frac{t}{T} \in [0,1]$.
We linearly interpolate the weights:
\[
\lambda_{\text{diag}}(t) = \lambda_{\text{diag}}^{\max}\cdot i_t,
\qquad
\lambda_{\text{mem}}(t) = \lambda_{\text{mem}}^{\max}\cdot (1-i_t),
\]
so that early rounds emphasize memory formation while later rounds emphasize diagnostic accuracy.
The shaped per-round reward is:
\vspace{-0.05in}
\[
\tilde r_t \;=\; \lambda_{\text{diag}}(t)\, r_t^{\text{diag}} + \lambda_{\text{mem}}(t)\, r_t^{\text{mem}}.
\]
\noindent\textbf{Round-wise advantage estimation.}
Because $\tilde r_t$ is non-stationary across $t$ under this schedule, trajectory-level normalization can cause later rounds to dominate
updates and make early-round gradients noisy.
We therefore compute advantages \emph{separately for each round position}.
For each round $t$, we sample a group of $K$ rollouts and obtain $\{\tilde r_t^{(k)}\}_{k=1}^{K}$.
We compute the group-relative advantage at round $t$:
\vspace{-0.05in}
\[
\hat A_t^{(k)} \;=\; \tilde r_t^{(k)} \;-\; \frac{1}{K}\sum_{j=1}^{K}\tilde r_t^{(j)}.
\]
\vspace{-0.05in}
We then optimize the policy with the standard clipped GRPO objective \cite{shao2024deepseekmath}, using $\hat A_t^{(k)}$ in place of
trajectory-level advantages.
\section{Experiments}
\label{sec:experiments}

We evaluate \ModelName{} in two complementary settings that isolate (i) \emph{static diagnostic competence}:
Standard Evaluation measures accuracy when each case is solved independently, without carrying experience across cases; (ii) \emph{continual on-the-job
improvement}: Long-Horizon Evaluation simulates deployment in a streaming environment, where the agent receives feedback after each case and must adapt over time via the memory mechanisms. More details on \ModelName{} implementation for both experiment settings is introduced in Appendix \ref{app:method_details}.

\subsection{Standard Evaluation Setting}
\label{sec:standard_eval}

This setting assesses \emph{static clinical reasoning ability}: Given a single case and a clinically plausible candidate set, can the model select the correct diagnosis? 

\subsubsection{Setup}
\noindent\textbf{Dataset.}
We use \textbf{MedCaseReasoning} \citep{wu2025medcasereasoning}, which contains challenging clinical cases paired with diagnosis labels.
We train on the official training split and report results on the test split, which emphasizes rare and complex diseases. 
We adopt a closed-set formulation to enable controlled and comparable evaluation across methods. By constraining the output space to a shared candidate set, we eliminate variability introduced by free-form generation and ensure that performance differences reflect reasoning quality rather than differences in answer expression. We fix the entire candidate-generation pipeline across methods: the distractor count $N$, sampling strategy, and random seeds are held constant. Importantly, we carefully design the distractor generation process to ensure high-quality and challenging candidate sets. Rather than sampling negatives uniformly, we construct clinically plausible and semantically similar distractors to the ground-truth diagnosis, requiring models to perform fine-grained reasoning rather than relying on superficial cues.
Thus, for each case, every method sees the same $\mathcal{Y}$ and is evaluated on the same closed-set decision. More details on the candidate-generation is described in Appendix \ref{app:candidates}.

\noindent\textbf{Evaluation Protocol.}
Each instance provides a patient case $x$ (demographics, symptoms) and a candidate set $\mathcal{Y}=\{y^{(1)},\dots,y^{(m)}\}$ containing the ground-truth diagnosis and $N=199$ clinically plausible distractors.
All methods must produce (i) a structured reasoning trace and (ii) a final prediction $\hat{y}\in\mathcal{Y}$, from which accuracy is computed. More details on evaluation setting is attached to Appendix \ref{app:eval}.

\noindent\textbf{Implementation.}
We use Qwen3-4b and Qwen3-8B \citep{qwen3} as the base policy model for \ModelName{} and train with the paradigm introduced in Section~\ref{sec:method_reward}. We additionally choose OLMo3-7B\citep{olmo2025olmo3}, LLaMa-3.1-8B\citep{llama}, and MedGemma-27B\citep{medgemma} as baseline models, spanning both general-domain and medical-domain settings as well as a range of model sizes. We compare against: \textbf{(i)} Zero-shot, \textbf{(ii)} Surpervised Finetuning Baselines, \textbf{(iii)} RL baselines, and
\textbf{(iv)} Memory augmented baselines. We fix the memory tool schema and maximum call limits to match \ModelName{} for fair comparison. More details on implementation is attached in Appendix \ref{app:baseline_impl}.

\begin{table}[tbh]
\centering
\small
\setlength{\tabcolsep}{12pt}
\renewcommand{\arraystretch}{1}
\definecolor{groupgray}{gray}{0.94}
\definecolor{darkgreen}{rgb}{0.0,0.5,0.0}
\begin{tabular}{l c c c}
\toprule
\textbf{Model} & \textbf{Accuracy (\%)} & \textbf{Macro-F1 (\%)} & \textbf{Std.} \\
\midrule

\rowcolor{groupgray}\multicolumn{4}{l}{\textit{Zeroshot}} \\

Qwen3-4b       & 43.2 & 40.0 & 0.025 \\
Qwen3-8b       & 45.1 & 41.9 & 0.025 \\
Qwen3-14B      & 50.4 & 42.9 & 0.024 \\
OLMo3-7B       & 47.1 & 43.9 & 0.025 \\
LLaMa-3.1-8B   & 42.1 & 39.8 & 0.025 \\
MedGemma-27B   & 54.7 & 51.1 & 0.023 \\

\addlinespace[4pt]
\rowcolor{groupgray}\multicolumn{4}{l}{\textit{Supervised Fine-Tuning (SFT)}} \\

Qwen3-4b (SFT) & 72.3 & 70.1 & 0.020 \\
Qwen3-8b (SFT) & 72.1 & 69.0 & 0.020 \\

\addlinespace[4pt]
\rowcolor{groupgray}\multicolumn{4}{l}{\textit{Reinforcement Learning (RL)}} \\

Qwen3-4b (RL-DiagnosticRewardOnly) & 74.1 & 69.5 & 0.025 \\
Qwen3-8b (RL-DiagnosticRewardOnly) & 77.3 & 72.3 & 0.023 \\

\addlinespace[4pt]
\rowcolor{groupgray}\multicolumn{4}{l}{\textit{Memory-Augmented}} \\

Qwen3-8b (ReAct+ShortTerm-Memory) & 42.5 & 38.3 & 0.025 \\
Qwen3-8b (SFT+ShortTerm-Memory)   & 56.3 & 50.9 & 0.023 \\

\midrule
\ModelName{} (Qwen-4b)
& 84.5 {\color{darkgreen}(+9.3\%)}
& 81.8 {\color{darkgreen}(+13.1\%)}
& 0.015 \\

\textbf{\ModelName{} (Qwen-8b)}
& \textbf{92.5 {\color{darkgreen}(+19.6\%)}}
& \textbf{86.8 {\color{darkgreen}(+20.1\%)}}
& 0.012 \\

\bottomrule
\end{tabular}

\caption{Performance comparison across methods under the standard evaluation setting.}
\label{tab:medcase_main}
\vspace{-0.2in}
\end{table}

\subsubsection{Results}
Table~\ref{tab:medcase_main} reports experiment results. We conclude three insights as following.

\noindent\textbf{Insight 1: Joint Reasoning–Memory Optimization Yields Large Gains.}
\ModelName{} consistently outperforms all baselines by a large margin. The Qwen-8B variant achieves 92.5\% accuracy and 86.81 macro-F1, surpassing the strongest baseline by +19.6\% and +20.1\%, respectively. Notably, similar improvements hold for the 4B model, suggesting that the gains are not solely due to scale but arise from the proposed joint optimization of reasoning and memory.

\noindent\textbf{Insight 2: Outcome-Only Reward Is Insufficient for Complex Diagnosis.}
While reinforcement learning improves over supervised fine-tuning, the gains remain modest. This indicates that optimizing only for final diagnostic correctness provides sparse and under-specified supervision. In contrast, \ModelName{} introduces structured rewards that guide both intermediate reasoning and memory usage, leading to substantially better performance on challenging clinical cases.

\noindent\textbf{Insight 3: Naive Memory Integration Can Hurt Performance.}
Memory-augmented baselines show that simply adding memory is not effective. ReAct with short-term memory drops to \textbf{42.5\%}, even below zeroshot performance, suggesting that unstructured retrieval introduces noise and disrupts reasoning. Even when combined with SFT, memory yields limited improvement, far below RL-based methods.

Overall, \ModelName{} substantially outperforms all baselines with results suggesting that jointly optimizing reasoning and memory is effective.

\subsection{Long-Horizon Evaluation Setting}
\label{sec:long_horizon}

This setting assesses \emph{on-the-job improvement}: Can the agent convert per-case feedback into reusable experience that improves future diagnoses over a stream?

\subsubsection{Setup}
\noindent\textbf{Dataset.}
We use \textbf{ER-Reason} \citep{mehandru2025er}, which contains heterogeneous emergency-room cases. We construct $\mathcal{Y}_t$ using the same distractor procedure as in Section~\ref{sec:standard_eval}, ensuring comparable decision difficulty across methods.

\noindent\textbf{Evaluation Protocol.}
We do \emph{not} train on ER-Reason. Instead, we create a single-pass stream $\{x_t\}_{t=1}^{T}$.
At each round $t$, the agent receives a case $x_t$ and a candidate set $\mathcal{Y}_t$, predicts $\hat{y}_t$, then receives feedback $f_t$
(e.g., the ground-truth label) before proceeding to $t{+}1$.
This protocol explicitly evaluates whether the agent improves as more feedback is observed. We design two evaluation metrics:
(1) \textbf{Final Accuracy}: the average top-1 accuracy computed over the entire interaction stream, reflecting the overall diagnostic performance after all learning and adaptation has taken place.
(2) \textbf{Improvement over $n$ turns}: 
$\Delta\text{Acc@}n=\text{Acc}(1{:}n)-\text{Acc}(1{:}n_0)$ with warm-up $n_0=10$, which measures the relative performance gain compared to an early-stage performance. More details on evaluation setting is attached to Appendix \ref{app:eval}.

\noindent\textbf{Implementation}
We select representative trained models from previous experiments. Model weights are frozen for all methods. We additionally compare against a strong general reasoning model GPT-5.2 \citep{gpt5} in the zero-shot and memory-augmented setting. More details on implementation is attached in Appendix \ref{app:baseline_impl}.

\subsubsection{Results}
Table~\ref{tab:ER-Reason_longhorizon} summarizes performance under the long-horizon streaming setting. We highlight three key findings from experiment results.

\begin{figure}[tbh]
    \centering
    \includegraphics[width=1\linewidth]{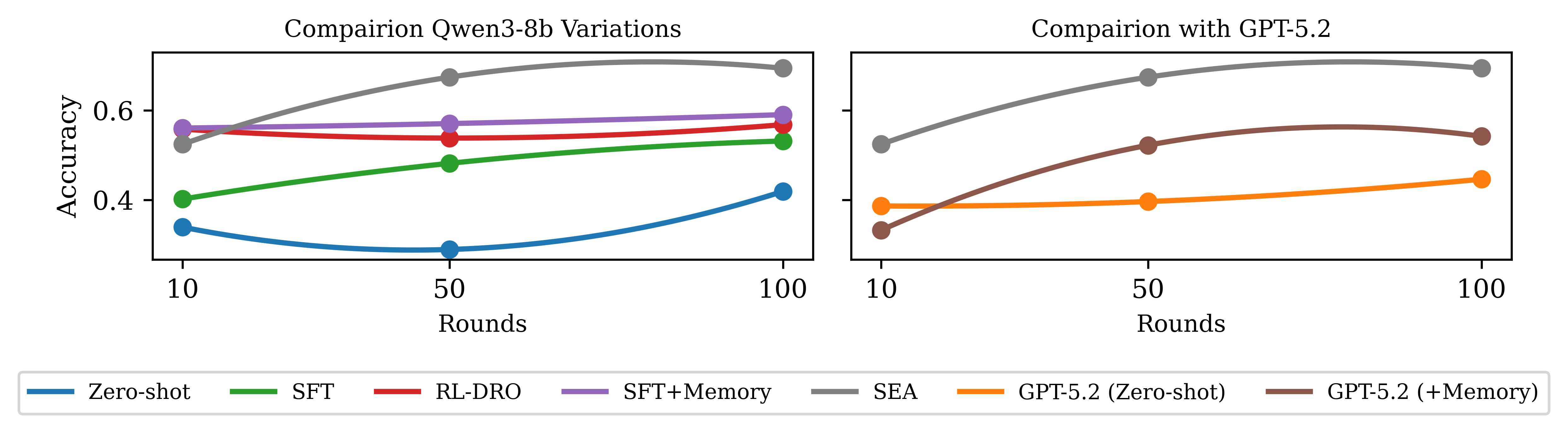}
    \caption{Accuracy trajectories from 10 to 100 rounds for representative methods.}
    \label{fig:accc_rounds}
\end{figure}

\begin{table}[tbh]
\centering
\small
\setlength{\tabcolsep}{6pt}
\renewcommand{\arraystretch}{1}
\definecolor{groupgray}{gray}{0.94}

\begin{tabular}{l c c c}
\toprule
\textbf{Method} & \textbf{Final Accuracy (\%)} & $\boldsymbol{\Delta}$\textbf{Acc@50 (\%)} & $\boldsymbol{\Delta}$\textbf{Acc@100 (\%)} \\
\midrule

\rowcolor{groupgray}\multicolumn{4}{l}{\textit{Zeroshot}} \\

Qwen-4b & 41.9 & -5.0 & +8.0 \\
Qwen-8b & 44.6 & +1.0 & +6.0 \\
GPT-5.2 & 53.2 & +8.0 & +13.0 \\

\addlinespace
\rowcolor{groupgray}\multicolumn{4}{l}{\textit{Trained Baselines}} \\

Qwen-8b (SFT) & 56.8 & -2.0 & +1.0 \\
Qwen-8b (RL-DiagnosticRewardOnly) & 59.1 & +1.0 & +3.0 \\

\addlinespace
\rowcolor{groupgray}\multicolumn{4}{l}{\textit{Memory-Augmented}} \\

Qwen-8b (SFT + Dual Memory) & 54.2 & +19.0 & +21.0 \\
GPT-5.2 (Zeroshot + Dual Memory) & 69.4 & +15.0 & +17.0 \\

\addlinespace
\midrule

SEA (Qwen-4b) & 68.5 & \textbf{+23.0} & +32.0 \\
\textbf{SEA (Qwen-8b)} & \textbf{72.1} & +20.0 & \textbf{+35.0} \\
\bottomrule
\end{tabular}
\caption{Final accuracy and improvement across different methods.}
\label{tab:ER-Reason_longhorizon}
\end{table}

\noindent\textbf{Insight 1: Memory alone enables adaptation but introduces a trade-off.}
Augmenting models with dual-memory significantly improves long-horizon adaptation. Both Qwen-8b (SFT + Dual Memory) and GPT-5.2 (Zeroshot + Dual Memory) show substantial gains ($+0.21$ and $+0.17$ at 100 rounds, respectively), confirming that external memory provides a mechanism for leveraging past feedback. However, this comes at a cost: Qwen-8b (SFT + Dual Memory) suffers a drop in final accuracy compared to its SFT counterpart (54\% vs.\ 56\%), suggesting that naive memory usage can introduce noise or retrieval errors that hurt overall decision quality.

\noindent\textbf{Insight 2: SEA achieves both strong final performance and sustained improvement.}
SEA outperforms baselines across both final accuracy and improvement metrics. In particular, SEA (Qwen-8b) achieves the best final accuracy (0.72) and the largest long-horizon gain ($\Delta$Acc@100 = $+0.35$), substantially exceeding both memory-augmented and trained baselines. Notably, SEA avoids the trade-off observed in naive memory systems: it improves continuously over time while also maintaining superior overall accuracy. 

\noindent\textbf{Insight 3: Dual-memory is a plug-and-play module for effective memory management.}
Our dual-memory design serves as an external, plug-and-play module that can be seamlessly integrated with different backbone models to enable structured memory management. Without any additional training, it equips models with the ability to store, retrieve, and consolidate experience over time. Augmenting GPT-5.2 with dual-memory improves final accuracy from 0.53 to 0.69 and increases long-horizon gain from $+0.13$ to $+0.17$ at 100 rounds. These results demonstrate that the benefits of our approach generalize across models, highlighting that effective adaptation can be achieved through a modular and reusable memory mechanism without modifying model parameters.

These results demonstrate that the ability to \emph{learn during deployment} is not an emergent property of standard LLMs or conventional training pipelines. 

\section{Discussion}
\label{sec:discussion}

\subsection{Expert Evaluation}
The central design of \ModelName{} is the dual-memory module, which enables the agent to induce reusable diagnostic rules from historical cases. While quantitative results suggest improved reasoning performance, it remains essential to assess whether the induced rules reflect clinically meaningful knowledge rather than superficial heuristics that improve task metrics. 

To gain further insight into the nature of these rules, we conduct a human evaluation with a professional physician. For each case, the evaluator is presented with the patient context, the model’s reasoning trace, and the induced rule, and is asked to assess them based on clinical validity and practical usefulness. Note that \textbf{this evaluation is intended to provide a qualitative perspective on the generated rules and to better understand how \ModelName{} leverages experience, rather than to serve as a definitive assessment of clinical quality}. We additionally provided the real trajectory from \ModelName{} in Appendix \ref{app:case}.

\noindent\textbf{Evaluation Dimensions.}
We evaluate the updated reasoning and induced rules along four dimensions:
(1) \emph{Clinical Correctness}, which measures factual accuracy and medical soundness;
(2) \emph{Case Relevance}, which assesses whether the reasoning and rules are properly grounded in patient-specific details;
(3) \emph{Rule Usefulness}, which captures the extent to which the induced rules are actionable and applicable in real diagnostic scenarios; and
(4) \emph{Overall Trust}, which reflects the physician's holistic confidence in the reasoning and rule quality.
Each dimension is rated on a 5-point Likert scale, where higher scores indicate better performance. More details are provided in Appendix~\ref{app:human_study}.

\noindent\textbf{Results.}
In the expert evaluation, \ModelName{} achieves an overall average score of 3.655 out of 5. The model demonstrates high \emph{overall trust} (4.2) and \emph{rule usefulness} (3.9), indicating that the induced rules are perceived as reliable and practically applicable in clinical settings. \emph{Clinical correctness} also remains high (3.8), suggesting that both the reasoning and induced rules are largely medically sound. Interestingly, \emph{case relevance} receives a relatively lower score (2.6). We emphasize that this behavior is expected and aligns with the design objective of \ModelName{}. The goal of rule induction is to abstract generalizable diagnostic principles from individual cases, rather than overfit to case-specific details. As a result, the induced rules intentionally prioritize cross-case applicability over tight coupling to a single instance.

% \begin{table}[t]
% \centering
% \small
% \setlength{\textfloatsep}{8pt plus 2pt minus 2pt}
% \setlength{\intextsep}{8pt plus 2pt minus 2pt}
% \setlength{\floatsep}{6pt plus 2pt minus 2pt}
% \renewcommand{\arraystretch}{1.05}
% \begin{tabular}{@{}l l c@{}}
% \toprule
% \textbf{Method} & \textbf{Ablated Module} & \textbf{Accuracy} \\
% \midrule
% Zero-shot & Training and Memory Modules & 0.4511 \\
% DiagnosticReward (w/o memory) & All Memory Modules & 0.7732 \\
% DiagnosticReward + ShortTerm-memory & Longterm Memory and Memory Reward & 0.8225 \\
% DiagnosticReward + LongTerm-Memory & Shortterm Mmory and Memory Reward & 0.7244 \\
% DiagnosticReward + Dual-Memory & Memory Reward & 0.4741 \\
% \midrule
% \textbf{\ModelName{}} & None (full model) & \textbf{0.9246} \\
% \bottomrule \\
% \end{tabular}
% \vspace{-0.2in}
% \caption{Ablation Study Results on MedCaseReasoning. \yue{for better readability: add $\Delta$ compared to SEA for each method}}
% \vspace{-0.25in}
% \label{tab:ablation_medcase}
% \end{table}

\subsection{Ablation Study}

\begin{table}[h]
\centering
\small
\setlength{\textfloatsep}{5pt plus 2pt minus 2pt}
\setlength{\intextsep}{5pt plus 2pt minus 2pt}
\setlength{\floatsep}{2pt plus 1pt minus 1pt}
\renewcommand{\arraystretch}{1.05}
\begin{tabular}{@{}l l c c@{}}
\toprule
\textbf{Method} & \textbf{Ablated Module} & \textbf{Acc.} & $\boldsymbol{\Delta}$ \\
\midrule
Zero-shot & Training \& Memory Modules & 0.45 & \cellcolor{darkred}-0.47 \\
DiagnosticReward (w/o memory) & All Memory Modules & 0.77 & \cellcolor{midred}-0.15 \\
DiagnosticReward+ShortTerm-memory & LongTerm Memory\&MemoryReward & 0.82 & \cellcolor{lightred}-0.10 \\
DiagnosticReward+LongTerm-Memory & ShortTerm Memory\&MemoryReward & 0.72 & \cellcolor{midred}-0.20 \\
DiagnosticReward+Dual-Memory & Memory Reward & 0.47 & \cellcolor{darkred}-0.45 \\
\midrule
\textbf{\ModelName{}} & None (full model) & \textbf{0.92} & --- \\
\bottomrule
\end{tabular}

\caption{Ablation study on MedCaseReasoning. $\Delta$ denotes performance change relative to the full model. Darker color indicates larger performance drop.}
\label{tab:ablation_medcase}
\end{table}

To identify which components contribute most to \ModelName{}'s gains, we perform controlled ablations on MedCaseReasoning. Table ~\ref{tab:ablation_medcase} reports the results. This result demonstrates that retaining recent case-specific information helps the model better align its reasoning with the current diagnostic context, leading to more accurate predictions. While long-term memory captures generalizable patterns across cases, it is less effective without mechanisms that ensure its relevance to the current instance. Combining short-term and long-term memory without the memory-management reward leads to a sharp performance drop to 0.4741, even below the zero-shot baseline. This indicates that unregulated memory interaction can introduce noise and conflicting signals, ultimately harming the reasoning process.
The full model achieves the best performance.
\vspace{-0.02in}
\section{Conclusion}
\vspace{-0.02in}
\label{sec:conclusion}

We propose \ModelName{}, a self-learning diagnostic agent with a cognitively inspired dual-memory module, trained via a reinforcement framework that jointly optimizes reasoning and memory management. Across two settings, \ModelName{} consistently outperforms strong baselines: It achieves 92.46\% accuracy on MedCaseReasoning with standard evaluation setting (+19.6\%) and attains the best final accuracy (0.7214) with the largest improvement (+0.35 $\Delta$Acc@100) on the ER-Reason dataset with long-horizon setting.

\section*{Acknowledgment}
This work used Delta GPU at NCSA through allocation [CIS240504] from the Advanced Cyberinfrastructure Coordination Ecosystem: Services \& Support (ACCESS) program, which is supported by U.S. National Science Foundation grants \#2138259, \#2138286, \#2138307, \#2137603, and \#2138296.

\bibliography{colm2026_conference}
\bibliographystyle{colm2026_conference}

\appendix
\newpage
\appendix
\onecolumn

\definecolor{codegreen}{rgb}{0,0.6,0}
\definecolor{codegray}{rgb}{0.5,0.5,0.5}
\definecolor{codepurple}{rgb}{0.58,0,0.82}
\definecolor{backcolour}{rgb}{0.95,0.95,0.92}
\lstdefinestyle{mystyle}{
    backgroundcolor=\color{backcolour},
    commentstyle=\color{codegreen},
    keywordstyle=\color{magenta},
    numberstyle=\tiny\color{codegray},
    stringstyle=\color{codepurple},
    basicstyle=\ttfamily\footnotesize,
    breakatwhitespace=false,
    breaklines=true,
    captionpos=b,
    keepspaces=true,
    % numbers=left,
    numbersep=5pt,
    showspaces=false,
    showstringspaces=false,
    showtabs=false,
    tabsize=2
}
\lstset{style=mystyle}

\section*{Appendix}

\section{Use of LLMs}
In this work, LLMs are used strictly for research support rather than as sources of substantive content. Their use falls into: (i) serving as the tested and trained model, and (ii) assisting with language refinement during paper writing. For writing support, we used GPT-5 solely to polish text (improving coherence and grammar) while all ideas, logic, results, and technical contributions originate from the authors. We note that all GPT-5 calls involving the data are made through an Azure-hosted OpenAI deployment, which provides data-handling safeguards for protected and access-controlled information.

\section{Limitations and Future Works}
\label{app:limitations}
Our study intentionally adopts a controlled closed-set diagnosis setting with immediate feedback, allowing us to isolate and rigorously evaluate the effect of memory-based adaptation independent of parameter updates. This design provides a clean testbed to demonstrate that substantial performance gains can be achieved through experience accumulation and memory updates alone, without modifying model weights. We emphasize that these choices are intentional design decisions to isolate the core research question: \emph{can experience alone, when properly structured and stored, drive meaningful improvement?} By controlling the output space and supervision signal, we create a clean and reproducible environment to rigorously study memory-based adaptation without confounding factors.
While real-world clinical environments may involve broader action spaces (e.g., test selection and evidence gathering), and strict safety requirements, we view our current setting as a necessary first step toward such complexity. It establishes a reliable foundation upon which these additional challenges can be systematically incorporated.

\section{Preliminaries}
\label{sec:prelim}

\begin{figure}[thbp]
    \centering
    \includegraphics[width=0.5\linewidth]{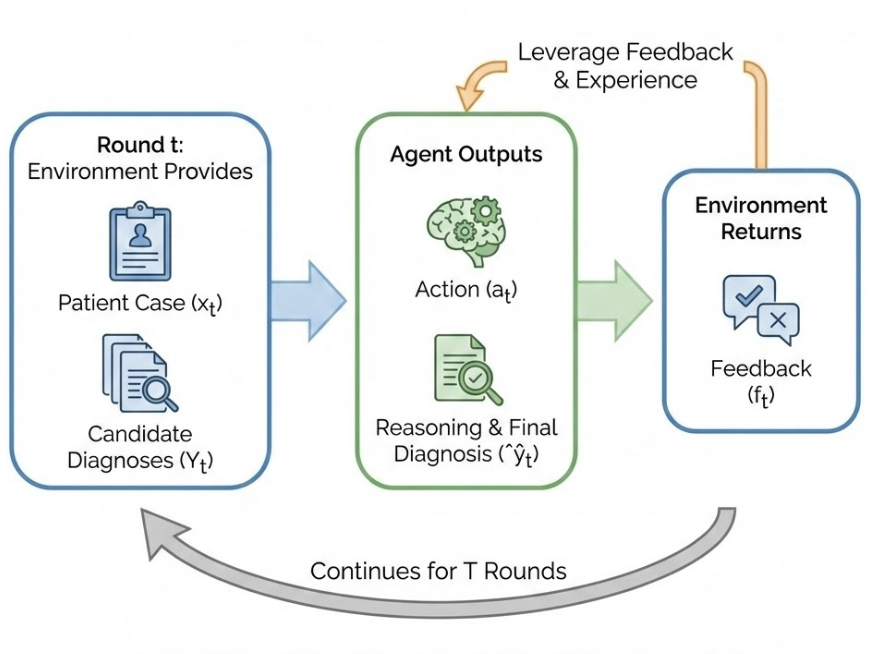}
    \caption{\textbf{Task setup:}
At each round $t$, the environment provides a patient case $x_t$ together with a candidate diagnosis set $\mathcal{Y}_t$.
The agent outputs an action $a_t$ consisting of a structured reasoning trace and a final prediction $\hat{y}_t\in\mathcal{Y}_t$.
The environment then returns feedback $f_t$ (e.g., correct/incorrect or graded), which the agent can leverage as experience for subsequent rounds.
This interaction repeats for $T$ rounds, modeling continual on-the-job improvement in a streaming setting.}
    \label{fig:tasksetup}
    \vspace{-0.1in}
\end{figure}

\noindent\textbf{Long-Horizon Setting Setup.}
We study long-horizon disease diagnosis as a sequential decision process motivated by \emph{on-the-job improvement} in deployment.
In real clinical settings, a diagnostic agent faces a stream of cases from a specific hospital and population, where prevalence, documentation style, and clinical practice can drift over time.
A fixed test-time reasoner may be competent on average yet brittle, repeatedly failing on recurring subtypes or missing subtle local cues. In practice, however, continuously retraining the base model during deployment is often infeasible due to privacy, safety, and operational constraints.
This motivates an agent that can \emph{learn from feedback and experience during use}, improving over the stream via a compact internal state (e.g., memory) rather than frequent model retraining.

At round $t \in \{1,\dots,T\}$, the environment provides a patient case
$x_t$ (e.g., demographics, symptoms, and clinical notes) together with a candidate diagnosis set
$\mathcal{Y}_t = \{y_t^{(1)}, \dots, y_t^{(m)}\}$.
The agent outputs an action $a_t$ that contains (i) a structured reasoning trace and (ii) a final predicted diagnosis
$\hat{y}_t \in \mathcal{Y}_t$ (or $\hat{y}_t \in \mathcal{Y}$ in the open-set setting).
After acting, the environment returns feedback $f_t$, which may include correctness, partial credit, or other evaluative signals.
The interaction then continues to the next round.

\noindent\textbf{On-the-job improvement objective.}
Unlike static test-time inference, our objective is continual improvement during deployment: the agent should become more accurate over the stream by incorporating feedback and experience without retraining on the target test stream.
Formally, let $\pi_\theta$ denote the policy parameterized by $\theta$. We aim to learn a policy that maximizes the expected cumulative utility
over the stream while enabling performance gains over time.
Crucially, the agent maintains an internal state $s_t$ across rounds, which summarizes what it has learned so far (e.g., via memory).

\noindent\textbf{Agentic Reinforcement Learning.}
Following \citet{zhang2025landscape}, we adopt agentic reinforcement learning to solve the task. We optimize the diagnosis agent as a policy model $\pi_\theta$ that generates a structured output $o_t$ (reasoning trace + final prediction) conditioned on the current case and the agent state:
$o_t \sim \pi_\theta(\cdot \mid x_t, \mathcal{Y}_t, s_t)$, where the predicted diagnosis $\hat{y}_t$ is parsed from $o_t$.
Each output is scored by a task-specific reward function $r_t = R(x_t, \mathcal{Y}_t, o_t, f_t)$, which may incorporate decision correctness, partial credit, format constraints, and other criteria.

\noindent\textbf{Group Relative Policy Optimization (GRPO).}
We train the policy with Group Relative Policy Optimization (GRPO) \cite{shao2024deepseekmath}, a PPO-style method that avoids explicit value-function learning by using \emph{within-group} comparisons among multiple sampled outputs for the same input.
At each training step for round $t$, we draw a group of $K$ candidate structured outputs $\{o_t^{(k)}\}_{k=1}^{K}$ from the current policy $\pi_\theta(\cdot\mid x_t,\mathcal{Y}_t,s_t)$ and score each candidate with our round-wise reward to obtain $\{r_t^{(k)}\}_{k=1}^{K}$.
Instead of learning a critic to predict returns, GRPO estimates an advantage signal by \emph{centering} each candidate reward by the group mean, so candidates that outperform their peers receive positive advantages and those that underperform receive negative advantages.
We then update $\pi_\theta$ by increasing the likelihood of higher-advantage outputs while applying (i) a clipped policy-ratio constraint to prevent overly large updates and (ii) a KL regularizer that keeps the updated policy close to a reference policy $\pi_{\text{ref}}$ for stability.

\section{Details on Candidate Label Selection}
\label{app:candidates}

A core challenge in constructing candidate diagnosis sets is avoiding \emph{trivial} distractors. If negative options are easily
distinguishable, models may succeed by exploiting superficial cues (e.g., matching the affected organ system) rather than performing
clinically meaningful discrimination. For example, for a lung-related presentation, the gold diagnosis might involve pulmonary inflammation,
while distractors may be unrelated conditions of the leg or heart; such mismatches make the task gameable and overestimate diagnostic ability.

To mitigate this issue, we design distractors to be \emph{clinically plausible} and \emph{semantically close} to the ground-truth label.
Rather than sampling negatives uniformly at random from a global label pool (e.g., \(\sim 800\) disease names), we adopt a \textbf{cluster-then-sample}
strategy: given the gold disease name, we retrieve a set of the most relevant diseases from the label list and sample distractors from this
neighborhood. Concretely, we use GPT-5.2 \cite{gpt5} to score candidate disease names by clinical relatedness to the gold label, optionally
grouping the label space into coarse clusters (e.g., by organ system, etiology, or phenotype similarity). We then draw \(N\) distractors from the top-ranked cluster(s), ensuring that negatives share overlapping symptoms, risk factors, or differential-diagnosis structure with the gold
label. This produces candidate sets where success requires fine-grained reasoning (e.g., distinguishing etiologies, time courses, or key
discriminative findings) instead of shallow lexical or organ-level matching.
Finally, to ensure fair comparison across methods, we fix the entire candidate-generation pipeline: the distractor pool, retrieval/scoring
procedure, distractor count \(N=199\), and random seeds are held constant, so every model is evaluated on identical candidate sets.

\section{\ModelName{} Details}
\label{app:method_details}
\subsection{Dual-Memory Module}
We implement the dual-memory module as an external tool, enabling plug-and-play integration with different backbone models without modifying their internal parameters. Specifically, the agent interacts with the memory module through structured tool calls, which support memory writing, retrieval, and consolidation during sequential decision-making.

The dual-memory structure consists of two complementary components: a \emph{short-term memory} that stores recent case-specific experiences, and a \emph{long-term memory} that maintains distilled, reusable knowledge (e.g., diagnostic rules). During inference, the agent retrieves relevant context, appends new experiences after feedback, manages memory, and consolidates abstract rules. We incorporate a system prompt explicitly instructs the agent on when and how to invoke memory operations. 

\begin{lstlisting}[language=Python, basicstyle=\ttfamily\small, breaklines=true]
{
    "type": "function",
    "function": {
        "name": self.name,
        "description": self.description,
        "parameters": {
            "type": "object",
            "properties": {
                "action": {
                    "type": "string",
                    "enum": ["list", "append", "pop", "consolidate"],
                    "description": (
                        "Operation to run. "
                        "`list` returns current short-term cases and long-term rules. "
                        "`append` adds a case record to short-term memory. "
                        "`pop` evicts cases at given indices from short-term memory. "
                        "`consolidate` adds distilled diagnostic rules to long-term memory."
                    ),
                },
                "case_record": {
                    "type": "object",
                    "description": (
                        "Case record to append (used with `append`). "
                        "Should contain keys like case_summary, diagnosis, feedback, reasoning."
                    ),
                    "properties": {
                        "case_summary": {
                            "type": "string",
                            "description": "Brief summary of the patient case.",
                        },
                        "diagnosis": {
                            "type": "string",
                            "description": "The diagnosis that was made.",
                        },
                        "feedback": {
                            "type": "string",
                            "description": (
                                "Whether the diagnosis was correct or incorrect, "
                                "and the ground truth."
                            ),
                        },
                    },
                },
                "indices": {
                    "type": "array",
                    "items": {"type": "integer"},
                    "description": (
                        "Indices of short-term cases to evict (used with `pop`)."
                    ),
                },
                "rules": {
                    "type": "array",
                    "items": {"type": "string"},
                    "description": (
                        "Diagnostic rules to add to long-term memory "
                        "(used with `consolidate`). "
                        "Each rule should be a concise, reusable statement "
                        "(e.g. symptom-disease associations)."
                    ),
                },
            },
            "required": ["action"],
        },
    },
}
\end{lstlisting}

\subsection{Training Details}
\label{sec:training_details}

\paragraph{Compute and infrastructure.}
Training is conducted on 8$\times$ NVIDIA A100 GPUs.
We build our training pipeline on top of the rLLMs infrastructure, which provides distributed rollout sampling, reward computation, and optimizer sharding for efficient policy optimization. The full training consumes about 48 GPU hours for Qwen3-8b base model.

\paragraph{Rollout and truncation.}
Each training example corresponds to one round $t$ with patient case $x_t$ and candidates $\mathcal{Y}_t$.
We cap the maximum number of generation steps and tool calls per round to 10 turns. A rollout terminates when the model emits a final diagnosis action $\hat{y}_t$. The prompt exceed the max prompt length will be truncate.

\paragraph{Training Parameters.}
We adopt Qwen3-4B and Qwen3-8b as the backbone and optimize it with a low learning rate ($1\text{e-}6$) under a PPO framework, using a relatively loose clipping threshold (0.28) and sequence-level loss aggregation. Training is conducted with moderate batch sizes (32 for training) and long context windows (up to 16k tokens for both prompt and response). Rollouts are generated using vLLM in asynchronous mode, with 8 samples per prompt during training (temperature 0.7) and deterministic evaluation settings (1 sample, temperature 0.6, top-$p$ 0.95) for validation. The system runs on a single node with 8 GPUs, incorporating efficiency optimizations such as gradient checkpointing, controlled GPU memory utilization (0.85), and a token limit of 32k per GPU for PPO. Additionally, stepwise advantage estimation is enabled in per-step mode to provide finer-grained credit assignment during training.

\paragraph{Implementation notes.}
We keep decoding settings fixed as default across training and evaluation to reduce confounding factors (temperature, max tokens, and stop conditions), and we checkpoint the best model based on validation accuracy.

\section{Automatic Evaluation Details}
\label{app:eval}

\subsection{Evaluation Prompts Templates}

\subsubsection{Standard Evaluation Setting}

\begin{tcolorbox}[breakable, colback=gray!5,colframe=black,title=\textbf{Standard Evaluation Prompt}]
\textbf{Role:} You are a clinical reasoning evaluator tasked with diagnosing a patient based strictly on provided evidence.

\vspace{0.3em}
\textbf{Task:}

Given a \textbf{Patient Profile} and a list of \textbf{Candidate Diseases} (each with descriptions), determine the \textbf{single most likely final diagnosis}.

\vspace{0.5em}
\textbf{Core Principles:}
\begin{itemize}[leftmargin=*]
\item This is a \textbf{closed-set diagnostic task}: the answer must be selected from the provided Candidate Diseases list.
\item Decisions must rely on \textbf{explicit evidence} from the Patient Profile.
\item Prioritize \textbf{clinical consistency, specificity, and minimal contradictions}.
\end{itemize}

\vspace{0.5em}
\textbf{Strict Rules (must follow):}
\begin{enumerate}[leftmargin=*]
\item Select \textbf{EXACTLY ONE} diagnosis from the Candidate Diseases list.
\item Do \textbf{NOT} invent diagnoses or use synonyms not appearing in the list.
\item Output the diagnosis name \textbf{EXACTLY as written}.
\item Use \textbf{ONLY} information from the Patient Profile:
\begin{itemize}
\item Do not assume missing symptoms, labs, or history.
\item Do not rely on external knowledge beyond the provided descriptions.
\end{itemize}
\item If multiple candidates are plausible, choose the one with:
\begin{itemize}
\item The \textbf{most specific and comprehensive match}.
\item The \textbf{fewest contradictions}.
\end{itemize}
\item Prefer diagnoses that explain \textbf{key distinguishing features} (e.g., critical symptoms, lab findings, temporal patterns).
\item Do \textbf{NOT} output multiple answers, uncertainty, or extra commentary.
\end{enumerate}

\vspace{0.5em}
\textbf{Reasoning Requirements:}
\begin{itemize}[leftmargin=*]
\item Cite \textbf{key evidence} from the Patient Profile (symptoms, history, labs, timeline).
\item Justify why the selected diagnosis \textbf{fits best}.
\item Optionally explain why close alternatives are less suitable.
\item Keep reasoning \textbf{concise, evidence-grounded, and non-speculative}.
\end{itemize}

\vspace{0.5em}
\textbf{Output Format (strict):}

Return \textbf{ONLY} a valid JSON object (no markdown, no extra text):
\begin{verbatim}
{
"reasoning": "Step-by-step evidence-based justification.",
"final_diagnosis": "EXACT disease name from the candidate list"
}
\end{verbatim}

\vspace{0.5em}
\textbf{Input:}

\textbf{Patient Profile:} \\
\texttt{\{current\_case\_prompt\}}

\vspace{0.3em}
\textbf{Candidate Diseases and Descriptions:} \\
\texttt{\{current\_choices\}}
\end{tcolorbox}

\begin{tcolorbox}[breakable, colback=gray!5,colframe=black,title=\textbf{Standard Evaluation Prompt Template (ReAct)}]

\textbf{Role:} You are a clinical reasoning evaluator. You solve diagnosis tasks via iterative reasoning and evidence grounding.

\vspace{0.4em}
\textbf{Task:}

Given a \textbf{Patient Profile} and \textbf{Candidate Diseases}, determine the \textbf{single most likely final diagnosis}.

\vspace{0.5em}
\textbf{Available Information:}
\begin{itemize}[leftmargin=*]
\item \textbf{Patient Profile}: Current case information (symptoms, history, labs).
\item \textbf{Candidate Diseases}: Closed-set diagnosis options with descriptions.
\end{itemize}

\vspace{0.5em}
\textbf{ReAct Reasoning Process:}
You must follow an iterative reasoning format:
\begin{itemize}[leftmargin=*]
\item \textbf{Thought}: Analyze current evidence and identify key clinical signals.
\item \textbf{Action}: Examine the patient profile and compare against candidate diseases.
\item \textbf{Observation}: Extract useful insights and match findings to candidate descriptions.
\end{itemize}

Iteratively repeat the sequence of Thought, Action, and Observation, where each cycle refines your understanding of the case and progressively grounds the diagnosis in evidence.

Then produce a final reasoning summary and diagnosis.

\vspace{0.5em}
\textbf{Strict Rules (must follow):}
\begin{enumerate}[leftmargin=*]
\item Select \textbf{EXACTLY ONE} diagnosis from the Candidate Diseases list.
\item Do \textbf{NOT} invent diagnoses or use synonyms not in the list.
\item Output the diagnosis name \textbf{EXACTLY as written}.
\item Use \textbf{ONLY} information from the Patient Profile.
\item Do \textbf{NOT} assume missing facts.
\item Choose the diagnosis with:
\begin{itemize}
\item strongest evidence match,
\item highest specificity,
\item minimal contradictions.
\end{itemize}
\end{enumerate}

\vspace{0.5em}
\textbf{Reasoning Requirements:}
\begin{itemize}[leftmargin=*]
\item Ground decisions in \textbf{explicit patient evidence}.
\item Prefer concise, structured, and evidence-based reasoning.
\end{itemize}

\vspace{0.5em}
\textbf{Output Format (strict):}

Return \textbf{ONLY} a valid JSON object (no markdown, no extra text):
\begin{verbatim}
{
"reasoning": "Condensed reasoning summary integrating patient evidence",
"final_diagnosis": "EXACT disease name from the candidate list"
}
\end{verbatim}

\vspace{0.5em}
\textbf{Input:}

\textbf{Patient Profile:} \\
\texttt{\{current\_case\_prompt\}}

\vspace{0.3em}
\textbf{Candidate Diseases and Descriptions:} \\
\texttt{\{current\_choices\}}

\end{tcolorbox}

\begin{tcolorbox}[breakable, colback=gray!5,colframe=black,title=\textbf{Standard Evaluation Prompt Template (Memory-Augmented)}]

\textbf{Role:} You are a clinical reasoning evaluator with access to external memory. You solve diagnosis tasks via iterative reasoning and evidence grounding.

\vspace{0.4em}
\textbf{Task:}

Given a \textbf{Patient Profile}, \textbf{Candidate Diseases}, and \textbf{Memory}, determine the \textbf{single most likely final diagnosis}.

\vspace{0.5em}
\textbf{Available Information:}
\begin{itemize}[leftmargin=*]
\item \textbf{Patient Profile}: Current case information (symptoms, history, labs).
\item \textbf{Candidate Diseases}: Closed-set diagnosis options with descriptions.
\item \textbf{Memory}: Past cases and history.
\end{itemize}

\vspace{0.5em}
\textbf{ReAct Reasoning Process:}
You must follow an iterative reasoning format:
\begin{itemize}[leftmargin=*]
\item \textbf{Thought}: Analyze current evidence and identify key clinical signals.
\item \textbf{Action}: Decide how to use memory (e.g., retrieve relevant cases or rules).
\item \textbf{Observation}: Extract useful insights from memory or the patient profile.
\end{itemize}

Iteratively repeat the sequence of Thought, Action, and Observation, where each cycle refines your understanding of the case and progressively grounds the diagnosis in evidence.

Then produce a final reasoning summary and diagnosis.

\vspace{0.5em}
\textbf{Memory Usage Guidelines:}
\begin{itemize}[leftmargin=*]
\item Use memory as \textbf{supporting evidence}, not as a substitute for the Patient Profile.
\item Prefer memory entries that match \textbf{key attributes} (symptoms, labs, progression).
\item \textbf{Do NOT override} patient evidence with memory if they conflict.
\item Long-term memory (rules) should guide abstraction; short-term memory (cases) should support analogy.
\end{itemize}

\vspace{0.5em}
\textbf{Strict Rules (must follow):}
\begin{enumerate}[leftmargin=*]
\item Select \textbf{EXACTLY ONE} diagnosis from the Candidate Diseases list.
\item Do \textbf{NOT} invent diagnoses or use synonyms not in the list.
\item Output the diagnosis name \textbf{EXACTLY as written}.
\item Use \textbf{ONLY} information from the Patient Profile and provided Memory.
\item Do \textbf{NOT} assume missing facts.
\item Choose the diagnosis with:
\begin{itemize}
\item strongest evidence match,
\item highest specificity,
\item minimal contradictions.
\end{itemize}
\end{enumerate}

\vspace{0.5em}
\textbf{Reasoning Requirements:}
\begin{itemize}[leftmargin=*]
\item Ground decisions in \textbf{explicit patient evidence}.
\item Incorporate \textbf{relevant memory} when helpful.
\item Prefer concise, structured, and evidence-based reasoning.
\end{itemize}

\vspace{0.5em}
\textbf{Output Format (strict):}

Return \textbf{ONLY} a valid JSON object (no markdown, no extra text):
\begin{verbatim}
{
"reasoning": "Condensed reasoning summary integrating patient evidence",
"final_diagnosis": "EXACT disease name from the candidate list"
}
\end{verbatim}

\vspace{0.5em}
\textbf{Input:}

\textbf{Patient Profile:} \\
\texttt{\{current\_case\_prompt\}}

\vspace{0.3em}
\textbf{Candidate Diseases and Descriptions:} \\
\texttt{\{current\_choices\}}

\vspace{0.3em}
\textbf{Memory:}
\texttt{\{short\_term\_memory\}}

\end{tcolorbox}

\subsubsection{Long-Horizon Evaluation Setting}

\begin{tcolorbox}[breakable, colback=gray!5,colframe=black,title=\textbf{Long-Horizon Prompt Template}]

\textbf{Role:} You are a clinical reasoning agent with access to memory, operating in a sequential evaluation setting.

\vspace{0.4em}
\textbf{Task:}
At round $t$, given the current \textbf{Patient Profile}, \textbf{Candidate Diseases}, and \textbf{Memory}, determine the \textbf{single most likely diagnosis}.

\vspace{0.5em}
\textbf{Setting:}
\begin{itemize}[leftmargin=*]
\item Cases arrive sequentially; each case is observed once.
\item After each prediction, you receive \textbf{feedback} (e.g., correct diagnosis).
\item You may update and use \textbf{memory} to improve future decisions.
\end{itemize}

\vspace{0.5em}
\textbf{Memory Usage Guidelines:}
\begin{itemize}[leftmargin=*]
\item Use memory as \textbf{supporting evidence}, not a replacement for patient data.
\item Do \textbf{NOT override} patient evidence with memory if they conflict.
\end{itemize}

\vspace{0.5em}
\textbf{Strict Rules:}
\begin{enumerate}[leftmargin=*]
\item Select \textbf{EXACTLY ONE} diagnosis from the Candidate Diseases.
\item Output the diagnosis name \textbf{EXACTLY as written}.
\item Base decisions on Patient Profile, optionally supported by Memory.
\item Do \textbf{NOT} assume missing information.
\end{enumerate}

\vspace{0.5em}
\textbf{Reasoning Requirements:}
\begin{itemize}[leftmargin=*]
\item Ground reasoning in explicit patient evidence.
\item Incorporate relevant memory when beneficial.
\item Clearly connect evidence (and memory, if used) to the diagnosis.
\end{itemize}

\vspace{0.5em}
\textbf{Output Format (strict):}
\begin{verbatim}
{
"reasoning": "Concise reasoning integrating patient
evidence and relevant memory.",
"final_diagnosis": "EXACT disease name"
}
\end{verbatim}

\vspace{0.3em}
\textbf{Memory:} \\
\texttt{$\{\mathcal{M}\}$}

\textbf{Input (Round $t$):}

\textbf{Patient Profile:} \\
\texttt{\{$x_t$\}}

\vspace{0.3em}
\textbf{Candidate Diseases:} \\
\texttt{$\{\mathcal{Y}_t\}$}

\end{tcolorbox}

\begin{tcolorbox}[breakable, colback=gray!5,colframe=black,title=\textbf{Long-Horizon Prompt Template (ReAct)}]

\textbf{Role:} You are a clinical reasoning agent with access to memory, operating in a sequential evaluation setting. You solve each diagnosis task through iterative reasoning grounded in patient evidence and relevant prior experience.

\vspace{0.4em}
\textbf{Task:}
At round $t$, given the current \textbf{Patient Profile}, \textbf{Candidate Diseases}, and \textbf{Memory}, determine the \textbf{single most likely diagnosis}.

\vspace{0.5em}
\textbf{Setting:}
\begin{itemize}[leftmargin=*]
\item Cases arrive sequentially; each case is observed once.
\item After each prediction, you receive \textbf{feedback} (e.g., correct diagnosis).
\item You may update and use \textbf{memory} to improve future decisions.
\end{itemize}

\vspace{0.5em}
\textbf{ReAct Reasoning Process:}
You must reason in an iterative format:
\begin{itemize}[leftmargin=*]
\item \textbf{Thought}: Analyze the current case and identify key clinical signals, uncertainties, or distinguishing findings.
\item \textbf{Action}: Decide whether to inspect the patient profile further, compare candidate diseases, or consult relevant memory.
\item \textbf{Observation}: Record the useful evidence obtained from the patient profile or memory, and refine the diagnostic hypothesis.
\end{itemize}

Iteratively repeat the sequence of \textbf{Thought}, \textbf{Action}, and \textbf{Observation} as needed, where each cycle refines your understanding of the case and progressively grounds the diagnosis in evidence.

\vspace{0.5em}
\textbf{Memory Usage Guidelines:}
\begin{itemize}[leftmargin=*]
\item Use memory as \textbf{supporting evidence}, not a replacement for patient data.
\item Do \textbf{NOT} override patient evidence with memory if they conflict.
\end{itemize}

\vspace{0.5em}
\textbf{Strict Rules:}
\begin{enumerate}[leftmargin=*]
\item Select \textbf{EXACTLY ONE} diagnosis from the Candidate Diseases.
\item Output the diagnosis name \textbf{EXACTLY as written}.
\item Base decisions on the Patient Profile, optionally supported by Memory.
\item Do \textbf{NOT} assume missing information.
\item Do \textbf{NOT} invent diagnoses or use synonyms not appearing in the candidate list.
\end{enumerate}

\vspace{0.5em}
\textbf{Reasoning Requirements:}
\begin{itemize}[leftmargin=*]
\item Ground reasoning in explicit patient evidence.
\item Incorporate relevant memory only when it provides useful support.
\item Clearly connect the selected diagnosis to the strongest supporting evidence.
\item Prefer concise, structured, and non-speculative reasoning.
\end{itemize}

\vspace{0.5em}
\textbf{Output Format (strict):}
\begin{verbatim}
{
"reasoning": "Concise reasoning integrating patient evidence 
and relevant memory.",
"final_diagnosis": "EXACT disease name"
}
\end{verbatim}

\vspace{0.3em}
\textbf{Memory:} \\
\texttt{$\{\mathcal{M}\}$}

\vspace{0.3em}
\textbf{Input (Round $t$):}

\textbf{Patient Profile:} \\
\texttt{\{$x_t$\}}

\vspace{0.3em}
\textbf{Candidate Diseases:} \\
\texttt{$\{\mathcal{Y}_t\}$}

\end{tcolorbox}

\subsection{Baseline Implementation Details}
\label{app:baseline_impl}

\subsubsection{Standard Evaluation Setting}
All baselines share the same input format, candidate set $\mathcal{Y}$, and output requirements (reasoning trace + final prediction), ensuring a controlled comparison. 

\textbf{Zero-shot} models directly perform inference using a unified prompting template without any parameter updates. 
\textbf{Supervised Fine-Tuning (SFT)} baselines are trained on the MedCaseReasoning training split using next-token prediction over reasoning traces and final answers, with the same data preprocessing and candidate sets as \ModelName{}. 
\textbf{RL baselines} are optimized with \emph{diagnostic reward only} from raw checkpoint, which provides scalar feedback based solely on final prediction correctness, without intermediate supervision or memory-related rewards. The training configuration (optimizer, batch size, rollout strategy) is kept identical to \ModelName{} to isolate the effect of reward design. 

For \textbf{memory-augmented baselines}, we equip models with the same memory tool interface but remove structured memory management. Specifically, \emph{ReAct+ShortTerm-Memory} uses tool calls to retrieve and append past cases but does not perform consolidation into long-term rules. \emph{SFT+ShortTerm-Memory} follows the same setup but leveraging the SFT-trained model. 

\subsubsection{Long-Horizon Evaluation Setting}
In the streaming setting, all methods operate under a single-pass protocol with frozen model parameters. 
\textbf{Zero-shot} and \textbf{trained baselines (SFT, RL)} perform inference independently at each step without retaining past information. 
\textbf{Memory-augmented baselines} extend these models with the dual-memory tool at inference time, allowing them to store case records after feedback and retrieve past experiences for future decisions. However, these baselines do not optimize memory usage.

For \textbf{GPT-5.2}, we evaluate both zero-shot and memory-augmented variants under the same protocol. The memory-augmented version uses the identical tool schema and system prompt as \ModelName{}, but without any additional training.

\subsubsection{Evaluation protocols}

All inputs are fixed before the evaluation starts. For each test case, we prepared 10 cases sampled from the training split as memory context, analogous to few-shot examples. No previously evaluated test instance is written into memory or made available to subsequent test instances. Thus, each test case is evaluated independently with respect to the test set. SEA is able to consolidate the rule from providing short-term memory and store in the long-term memory, but the content in the long-term memory is not carried into the next test case. SEA and other memory-augmented baselines received the same short-term memory. In the ER-Reason long-horizon setting, cases are processed sequentially and feedback is provided after each prediction, allowing memory updates across rounds

\begin{table*}[h]
\centering
\caption{Comparison of standard and long-horizon evaluation settings.}
\label{tab:evaluation_settings}
\renewcommand{\arraystretch}{1.25}
\setlength{\tabcolsep}{8pt}

\begin{tabularx}{\textwidth}{
    >{\raggedright\arraybackslash}p{3.4cm}
    >{\raggedright\arraybackslash}X
    >{\raggedright\arraybackslash}X
}
\toprule
\textbf{Evaluation aspect}
&
\textbf{Standard evaluation}
&
\textbf{Long-horizon evaluation}
\\
\midrule

Dataset
&
MedCaseReasoning
&
ER-Reason
\\

Training on the target dataset
&
Yes, using its training split
&
No
\\

Test-case organization
&
Cases are evaluated independently
&
Cases arrive sequentially as a stream
\\

Feedback during evaluation
&
No
&
Yes, after each round
\\

Memory update during evaluation
&
No feedback-based self-learning
&
Yes, previous experience is incorporated
\\

Primary purpose
&
Evaluate static diagnostic competence
&
Evaluate OOD sequential adaptation and experience reuse
\\

\bottomrule
\end{tabularx}
\end{table*}

\section{Expert Study Details}
\label{app:human_study}

We conduct invite professional physician to assess the quality of the updated reasoning and induced rules along four dimensions. Annotator is asked to rate each dimension on a 5-point Likert scale, with higher scores indicating better performance.

\vspace{4pt}
\noindent\textbf{Evaluation Criteria.}

\begin{itemize}[leftmargin=*]

\item \textbf{Clinical Correctness.}
\emph{Does the updated reasoning and rules remain medically accurate and free of factual errors?}
\begin{itemize}
    \item 5: No factual errors; medically sound throughout
    \item 4: Very minor factual imprecision with negligible impact
    \item 3: Noticeable factual issue that could alter interpretation
    \item 2: Major factual error that undermines clinical validity
    \item 1: Severely incorrect or unsafe factual content
\end{itemize}

\item \textbf{Case Relevance.}
\emph{Does the reasoning appropriately incorporate case-specific details?}
\begin{itemize}
    \item 5: Fully grounded in key case details and diagnostic context
    \item 4: Mostly grounded with minor omissions or weak links
    \item 3: Partially grounded; several important details underused
    \item 2: Weakly grounded; reasoning is largely generic or misaligned
    \item 1: Not grounded in the case; off-target reasoning
\end{itemize}

\item \textbf{Rule Usefulness.}
\emph{Is the induced rule practically useful and likely to be applied?}
\begin{itemize}
    \item 5: Strongly like and would readily apply this rule
    \item 4: Generally like and would likely use in most cases
    \item 3: Neutral; might use in some situations
    \item 2: Dislike; unlikely to use except as a last resort
    \item 1: Strongly dislike and would not use this rule
\end{itemize}

\item \textbf{Overall Trust.}
\emph{How reliable is the updated reasoning and rule overall?}
\begin{itemize}
    \item 5: Very high trust; can be adopted with confidence
    \item 4: High trust; acceptable with minor caveats
    \item 3: Moderate trust; requires targeted review
    \item 2: Low trust; substantial concerns remain
    \item 1: Very low trust; should not be adopted
\end{itemize}

\end{itemize}

\vspace{4pt}
\noindent\textbf{Annotation Guidelines.}
Annotator is instructed to consider both the updated reasoning trace and the induced rules jointly. Clinical correctness should be judged strictly based on medical validity, while case relevance emphasizes grounding in the specific patient context. Rule usefulness captures practical applicability, and overall trust reflects a holistic judgment combining all aspects.

\noindent\textbf{Annotator Profile.}
Annotator is an internal medicine resident with a strong interdisciplinary background in clinical medicine, biostatistics, and oncology research. Recipient of multiple academic distinctions, including national and institutional scholarships.

% \section{Case Study}
% \label{app:case}
% We present 15 trajectory from \ModelName{} under the long-horizon setting in the supplementary materials to illustrate how the agent adapts its reasoning and memory over time to resolve complex diagnostic scenarios.

\end{document}